\documentclass[conference]{IEEEtran}
\IEEEoverridecommandlockouts
\usepackage{cite}
\usepackage{amsmath,amssymb,amsfonts}
\usepackage{algorithmic}
\usepackage{graphicx}
\usepackage{textcomp}
\usepackage{xcolor}
\usepackage{booktabs}
\usepackage{algorithm}
\usepackage{multirow}
\usepackage{bm}
\def\BibTeX{{\rm B\kern-.05em{\sc i\kern-.025em b}\kern-.08em
    T\kern-.1667em\lower.7ex\hbox{E}\kern-.125emX}}

\usepackage{float}
\usepackage{colortbl}

\def\orange#1{\textcolor{orange}{#1}}

\definecolor{DeepBlue}{RGB}{0, 50, 150}

\def\eg{\emph{e.g.}}

\def\ie{\emph{i.e.}}

\definecolor{mygray}{gray}{.9}

\usepackage[pagebackref,breaklinks,colorlinks]{hyperref}

\usepackage[capitalize]{cleveref}
\crefname{section}{Sec.}{Secs.}
\Crefname{section}{Section}{Sections}
\Crefname{table}{Table}{Tables}
\crefname{table}{Tab.}{Tabs.}

\begin{document}

\title{HRGR: Enhancing Image Manipulation Detection via Hierarchical Region-aware Graph Reasoning}


\author{\thanks{This work is supported in part by the National Natural Science Foundation of China (No.62402464), Shandong Provincial Natural Science Foundation (No.ZR2024QF035), and China Postdoctoral Science Foundation (No.2021TQ0314; No.2021M703036).}
Xudong Wang$^1$, Jiaran Zhou$^1$, Huiyu Zhou$^2$, Junyu Dong$^1$, Yuezun Li$^{1,\dagger}$\thanks{$\dagger$ Corresponding author} \\
$^1$ School of Computer Science and Technology, Ocean University of China \\
$^2$ School of Computing and Mathematical Sciences, University of Leicester}

\maketitle

\begin{abstract}
    Image manipulation detection is to identify the authenticity of each pixel in images. One typical approach to uncover manipulation traces is to model image correlations. 
    The previous methods commonly adopt the grids, which are fixed-size squares, as graph nodes to model correlations. However, these grids, being independent of image content, struggle to retain local content coherence, resulting in imprecise detection.
    To address this issue, we describe a new method named Hierarchical Region-aware Graph Reasoning (HRGR) to enhance image manipulation detection. Unlike existing grid-based methods, we model image correlations based on content-coherence feature regions with irregular shapes, generated by a novel Differentiable Feature Partition strategy. Then we construct a Hierarchical Region-aware Graph based on these regions within and across different feature layers. Subsequently, we describe a structural-agnostic graph reasoning strategy tailored for our graph to enhance the representation of nodes. 
    Our method is fully differentiable and can seamlessly integrate into mainstream networks in an end-to-end manner, without requiring additional supervision.
    Extensive experiments demonstrate the effectiveness of our method in image manipulation detection, exhibiting its great potential as a plug-and-play component for existing architectures.
    Codes and models are available at \url{https://github.com/OUC-VAS/HRGR-IMD}.
\end{abstract}


\section{Introduction}
Image manipulation involves various techniques such as copy-move, splicing, and object removal that can subtly alter the primary content of images. As image editing tools and AI technology continue to advance, image manipulation has become effortless, leading to significant societal issues such as misinformation spread, ethical dilemmas, and potential legal problems \cite{recent}. 

\begin{figure}[!t]
    \centering
    \vspace{0.3cm}
    \includegraphics[width=1\linewidth]{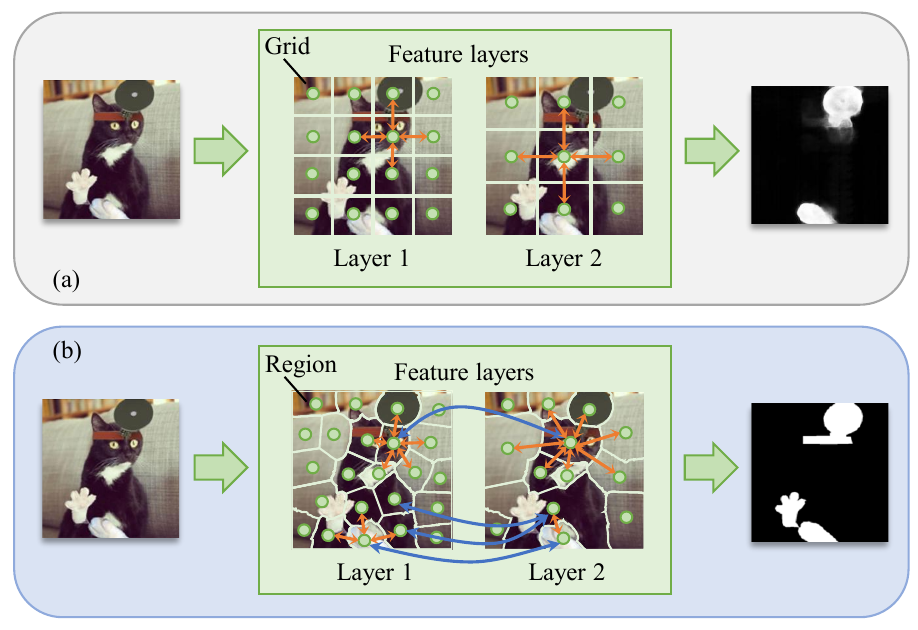}
    \vspace{-0.6cm}
    \caption{\small The difference between (a) previous methods and (b) ours. Instead of using regular grids in previous methods, we model image correlations across different scales (\orange{orange} and {\textcolor{DeepBlue}{blue}} arrows) using connected local content-coherent feature regions. }
    \label{fig:concept}
    \vspace{-0.5cm}
\end{figure}

To combat this, countermeasures have been proposed and gained increasing attention in recent years. Deep learning-based methods have emerged as the dominant method for detecting manipulations \eg, \cite{ERMPC,PSCC,ObjectFormer,CAT-Net,ACBG,TruFor},. Leveraging the impressive learning capabilities of deep learning models. Based on deep neural networks, feature enhancement strategies have been explored to further emphasize manipulation traces. 
Particularly, exploiting image correlations is a recent feature enhancement solution that has proven effective in enhancing detection ability \cite{ERMPC,PSCC}. The intuition is that manipulated regions can be better perceived when compared to pristine regions. To capture the correlations, these methods usually employ the mechanism of either attention mechanism \cite{PSCC,ObjectFormer} or graphs \cite{HGCN,ERMPC}. For example, PSCC-Net \cite{PSCC} proposed a spatial-channel correlation module to capture the correlations using attention. {ObjectFormer \cite{ObjectFormer} learned the correlation between intermediate features in objects and patches through cross-attention. HGCN \cite{HGCN} and ERMPC \cite{ERMPC} sought the relationship between pristine and manipulated regions by building a graph structure. }

\textbf{Despite their promising results, these methods share a common limitation: They all rely on the grid-based modeling of correlations.} It is important to note that these grids are regular squares with fixed sizes that are independent of the content. Typically, these methods partition the features into grids of size $n \times n$ and encapsulate the features within each grid as respective units for correlation modeling (see Fig.~\ref{fig:concept} (top)). However, as manipulations usually happen on the semantic-integrity objects (\eg, copy-move/splicing a whole person), the use of grids can hardly retain the local content coherence. This limitation can potentially hinder their representation and subsequently restrict the extraction of manipulation traces (\eg, leading to ambiguous detection boundaries (see Fig.~\ref{fig:concept} (top)).)




To overcome the limitation, we introduce a novel framework called {\em Hierarchical Region-aware Graph Reasoning (HRGR)} to enhance image manipulation detection. Instead of using regular grids to model correlations, we propose a Hierarchical Region-aware Graph that captures the correlations based on content-coherent \textbf{feature regions} (see Fig.~\ref{fig:concept} (bottom)). 


\begin{figure*}[!t]
    \centering
    \includegraphics[width=1\linewidth]{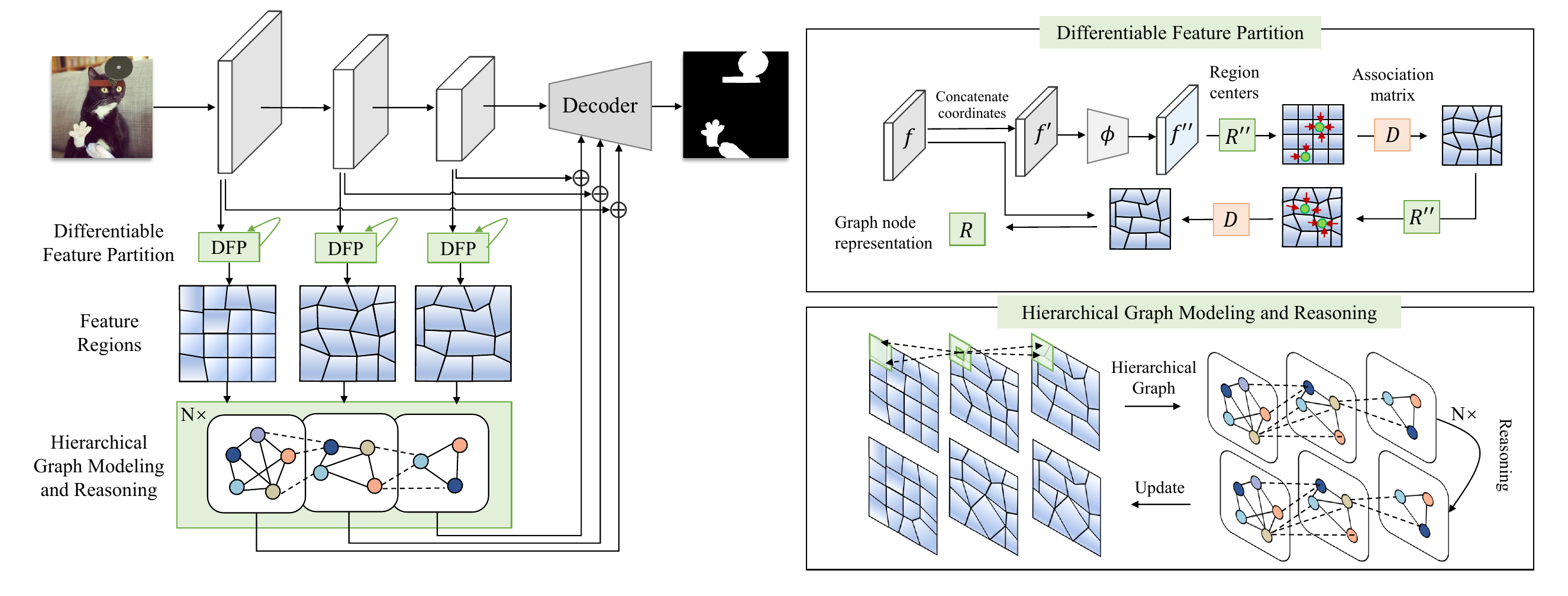}
    \vspace{-0.9cm}
    \caption{\small Overview of our method. We first extract the feature regions on feature layers from the encoder using a Differentiable Feature Partition (DFP) strategy. Then we build the hierarchical region-aware graph over all features and perform graph reasoning to update the features. The process of Differentiable Feature Partition and Hierarchical Graph Reasoning and Reasoning are shown in the right part.}
    \label{fig:overview}
    \vspace{-0.3cm}
\end{figure*}

To achieve this, we first describe a Differentiable Feature Partition strategy that divides the features into non-overlapped coherent regions while maintaining differentiability for end-to-end utilization. These feature regions are subsequently encapsulated as graph nodes using a Weighted Feature Aggregation strategy. In contrast to previous methods that typically construct graphs on single feature layers, we propose a Hierarchical Graph Modeling strategy to construct a graph within the single feature layers as well as across different feature layers. This is motivated by that the same manipulated instance at different scales should exhibit consistent manipulation traces. By modeling across different feature layers, we can identify this relation, thus further improving the identification of manipulation traces. After modeling the correlations, we describe a Structural-agnostic Graph Reasoning strategy to refine the representation of graph nodes, which are then mapped back to feature layers for feature enhancement. The overview of our method is illustrated in Fig.~\ref{fig:overview} (left).

It is noteworthy that our method is designed as a plug-and-play framework that can seamlessly integrate into mainstream networks without requiring additional supervision. 
Experiments are conducted on commonly used datasets, and the results demonstrate the efficacy of the proposed method.

The contribution of this paper can be summarized as follows: \textbf{1)} To the best of our knowledge, we are the first to explore image correlations based on content-coherent irregular feature regions rather than regular grids, facilitated by a novel Differentiable Feature Partition strategy. \textbf{2)} Leveraging the feature regions, we describe a new Hierarchical Region-aware Graph Reasoning strategy that seeks correlations within the same feature layer and across different feature layers. \textbf{3)} Our framework is fully differentiable and can seamlessly integrate into mainstream networks in an end-to-end manner without requiring additional supervision. 

\vspace{-0.2cm}
\section{Method}

\subsection{Hierarchical Region-aware Graph}
\label{sec:hrg}
With feature maps, the gist of constructing this graph lies in two main steps: 1) formulating graph nodes and 2) modeling the correlations among these graph nodes. To ensure the end-to-end fashion of our method, accomplishing these two steps is not trivial. In the following, we first outline the overview of our method and then elaborate on the details of each step.




\subsubsection{Differentiable Feature Partition} 
\label{sec:dfp}
To formulate the graph nodes, we propose a new Differentiable Feature Partition (DFP) strategy to create feature regions. This strategy can partition the features into several non-overlapped and content-coherent regions in an unsupervised way, and can be integrated into the training in an end-to-end fashion.

Intuitively, the conventional clustering algorithms (\eg, K-means) can be applied to feature layers by measuring the similarity among feature elements. However, this algorithm suffers from two difficulties. One is that local connectivity can not be guaranteed and the other is the clustering process is not differentiable. Inspired by the superpixel extraction methods \cite{SpixelFCN,SSN} and etc., we construct a soft feature element-region association matrix to make the clustering process differentiable.   

Denote a feature map from the encoder $\mathcal{E}$ as $f \in \mathbb{R}^{h \times w \times c}$, where $h,w,c$ are the corresponding number of feature height, width, and channel. To ensure region connectivity, we append the position (\ie, x- and y- coordinates) of each feature element on its channel dimension as $f' \in \mathbb{R}^{h \times w \times c'},c'=c+2$. 
Note that the time cost in clustering is proportional to feature dimensions. To reduce the computational overhead while refining the features, we design a convolutional block $\phi$ that transforms the features $f'' = \phi (f'), f'' \in \mathbb{R}^{h \times w \times c''} (c'' < c')$.
Based on $f''$, we initialize the region centers. Denote the number of regions as $m$. We evenly divide $f''$ into $m$ grids, and average the features inside each grid as the initial representation of region centers. Let the region centers be a matrix $\mathbf{R}'' = [ r''_1; ...; r''_m ] \in \mathbb{R}^{m \times c''}$. Each region center $r''_i$ has a size of $1 \times c''$. For clustering, we build the feature element-region association matrix to store the probability of assigning a feature element to a region, denoted as $\mathbf{D} \in \mathbb{R}^{n \times m}, n = h \times w$.

For the $i$-th feature element $e''_i \in f''$, we calculate the Euclidean distance between it and other region centers in $\mathbf{R}''$ and perform a softmax operation to normalize the probability, as 
\begin{equation}
\begin{aligned}
    \mathbf{D}(i, j) = \frac{{\exp}^{-\lVert e''_i - r''_j \rVert^{2}}}{\sum^{m}_{j=1} {\exp}^{-\lVert e''_i - r''_j \rVert^{2}}}.
\end{aligned}
\label{eq:prob}
\end{equation}

After obtaining the matrix $\mathbf{D}$, the intuitive way of obtaining regions is to assign each element to its nearest center, as
\begin{equation}
\begin{aligned}
    \mathbf{D}(i, j) = 
    \left \{
        \begin{array}{cl}
            1 & \text{if} \quad j = \mathop{\arg\max}\limits_{j \in \{1,...,m \}} (\mathbf{D}(i,j)), \\
            0 & \text{otherwise}.
        \end{array}
    \right.
\end{aligned}
\label{eq:assign}
\end{equation}
Then the region center $r''_j \in \mathbf{R}''$ can be updated by 
\begin{equation}
\begin{aligned}
    r''_j = \frac{\sum^{n}_{i=1} e''_i \cdot  \mathbf{D}(i, j) }{\sum^{n}_{i=1} \mathbf{D}(i, j)}.
\end{aligned}
\label{eq:center}
\end{equation}

However, this assignment of Eq.~\eqref{eq:assign} is not differentiable due to the \texttt{argmax} operation, which thus can not be incorporated into our model.

\smallskip
\textit{\includegraphics[width=0.3cm]{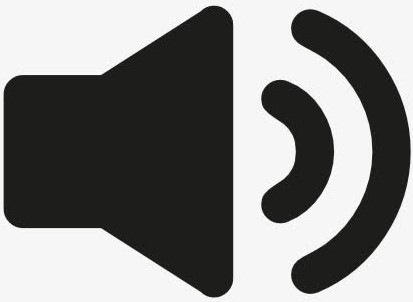}: ``How to make this process differentiable?''}
\smallskip

To address this issue, we use soft assignments, \ie, an element can be assigned to all regions with various probabilities. Specifically, we abandon Eq.~\eqref{eq:assign} and update the region centers using the probabilities in matrix $\mathbf{D}$, directly following Eq.~\eqref{eq:center}. The update process can be formulated as the operations between matrices as
\begin{equation}
\begin{aligned}
    & \mathbf{F}'' \in \mathbb{R}^{n \times c''} \leftarrow \text{Reshape}({f''} \in \mathbb{R}^{h \times w \times c''}), \\
    & \mathbf{R}'' = \frac{1}{\mathbf{N}_{\rm D}} \circ (\mathbf{D}^{\top} \mathbf{F}''),
\end{aligned}
\label{eq:center_m}
\end{equation}
where $\mathbf{N}_{\rm D} = [\sum^{n}_{i=1} \mathbf{D}(i, 1),...,\sum^{n}_{i=1} \mathbf{D}(i, m)]$ is the summation of each column of $\mathbf{D}$, $\circ$ denotes the element-wise product.

Note the feature regions are updated iteratively. Denote the current iteration as $t$. Once the region centers $\mathbf{R}''_{t}$ are updated, we can re-calculate the matrix $\mathbf{D}_{t+1}$ and then update $\mathbf{R}''_{t+1}$. This process is repeated until the maximum iteration $T$ is reached. The overview of this process is shown in Fig.~\ref{fig:overview} (right-top). 

\subsubsection{Weighted Feature Aggregation for Node Representation} 
\label{sec:nr}
After obtaining the feature regions, we encapsulate the feature elements inside regions as the representation of graph nodes using a Weighted Feature Aggregation strategy.
Denote the matrix of node representations as $\mathbf{R} = [ r_i;...;r_m ] \in \mathbb{R}^{m \times c}$. 
{The encapsulation process can be defined with similar formulas as Eq.~\eqref{eq:center_m}, with the difference that here the encapsulation is applied to the original encoder features $f$, as}

\begin{equation}
\begin{aligned}
    & \mathbf{F} \in \mathbb{R}^{n \times c} \leftarrow \text{Reshape}({f} \in \mathbb{R}^{h \times w \times c}), \\
    & \mathbf{R} = \frac{1}{\mathbf{N}_{\rm D}} \circ (\mathbf{D}^{\top} \mathbf{F}).
\end{aligned}
\label{eq:center_f}
\end{equation}

\subsubsection{Hierarchical Graph Modeling}
\label{sec:hgm}
\begin{figure}[!t]
    \centering
    \vspace{-0.35cm}
    \includegraphics[width=1\linewidth]{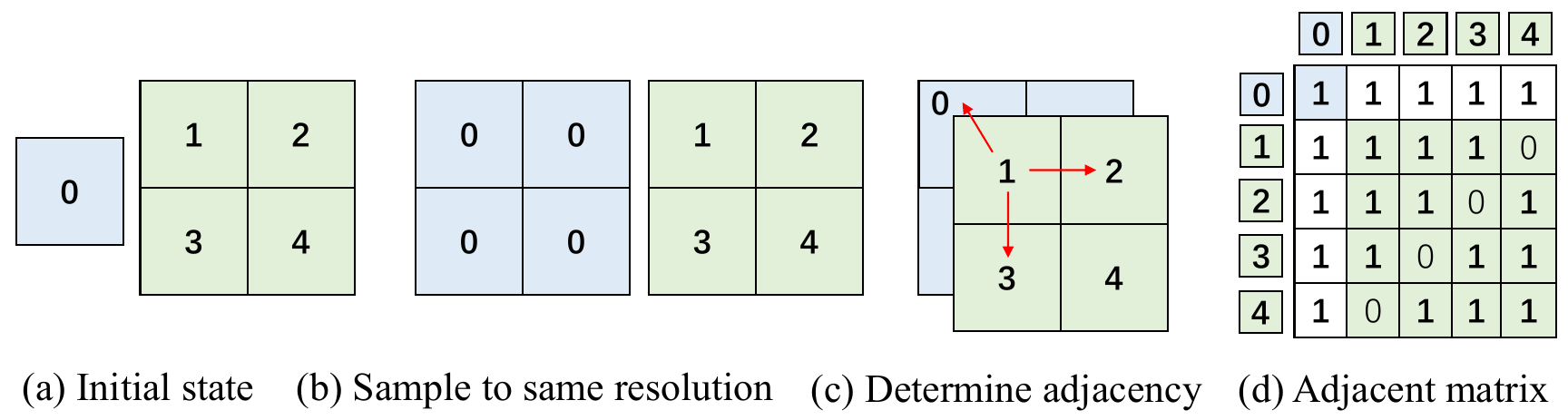}    
    \vspace{-0.6cm}
    \caption{Illustration of adjacent matrix $\mathbf{A}$. We build a hierarchical graph that establishes adjacency relationships within the same layer (intra-layer, orange arrows in Fig.~\ref{fig:concept} (b)) and also across different layers (inter-layer, blue arrows in Fig.~\ref{fig:concept} (b)), resulting in a single adjacent matrix $\mathbf{A}$. For simplicity, we use an example with two layers (highlighted in blue with one node and green with four nodes, see Fig~\ref{fig:relationships} (a)). First, we resample them to the same resolution (see Fig~\ref{fig:relationships} (b)). Then we stack them along channel dimension and apply the 3D sliding window strategy to obtain an adjacency matrix (see Fig~\ref{fig:relationships} (c,d)). }
    \vspace{-0.5cm}
    \label{fig:relationships}
\end{figure}

With the obtained graph nodes, we propose a Hierarchical Graph Modeling strategy to construct a graph on feature regions within the same and across different feature layers. 
Denote a set of feature maps from different layers as $\{f^1,...,f^k\}$. We first obtain the feature element-region matrix for each feature as $\{\mathbf{D}^1,...,\mathbf{D}^k\}$ and their corresponding node representations as $\{\mathbf{R}^1,...,\mathbf{R}^k\}$. Denote the adjacent matrix for this graph as $\mathbf{A} \in \mathbb{R}^{M \times M}$, where $M$ is the total number of nodes, \ie, $M = \sum^{k}_{i=1} | \mathbf{R}^i |$ (see Fig.~\ref{fig:relationships} for illustration).


\smallskip
\textit{\includegraphics[width=0.3cm]{figures/R.jpg}: ``How to seek the hierarchical neighborhood relationships?''}
\smallskip

Since the regions corresponding to nodes are irregular, modeling the relationship is more challenging than the conventional grid-based methods. 
As such, we describe a 3D sliding-window based strategy to seek the correlations. 
It is noteworthy that in this step, we require an assignment of feature elements to regions. Therefore, we reuse Eq.~\eqref{eq:assign} to divide each feature map into non-overlapped regions and assign each region a unique index from $\{1,...,M\}$. Then we can create the region index maps $\{J^1,...,J^k\}$ corresponding to $\{f^1,...,f^k\}$. We resize each index map to the same spatial size and concatenate them as a matrix $\mathbf{J} \in \mathbb{R}^{h \times w \times k}$. Then we perform the 3D sliding window with a size of $3 \times 3 \times 3$ in order. The criterion is that $\mathbf{A}(\mathbf{J}(p_0),\mathbf{J}(p_0+p_n)) = 1$, {if $\mathbf{J}(p_0) \neq \mathbf{J}(p_0+p_n)$, where $p_0$ is the position of the window center and $p_n$ is the offset within window.}
To accelerate this process, we formulate it into CUDA code.


Besides the adjacent matrix $\mathbf{A}$,
we construct the node representation matrix for all regions.
Since $\mathbf{R}^1,...,\mathbf{R}^k$ has different channel numbers, we perform a linear transform for each matrix to uniform their channel number as $C$. Denote the node representation matrix as $\mathbf{R}_{\rm G} \in \mathbb{R}^{M \times C}$, which is obtained by
\begin{equation}
    \begin{aligned}
        \mathbf{R}_{\rm G} & = [\mathbf{R}^{1}_{\rm G};...;\mathbf{R}^{k}_{\rm G}] = [\mathbf{R}^1 \mathbf{W}^1;...;\mathbf{R}^k \mathbf{W}^k],
    \end{aligned}
\end{equation}
where $\mathbf{W}^1,...,\mathbf{W}^k$ are learnable linear matrics.


\subsection{Structure-agnostic Graph Reasoning}
\label{sec:sgr}

In the design of our method, the feature regions can be dynamically updated during training, which subsequently results in different graph structures. 
To process the dynamic graphs, we adapt structure-agnostic graph reasoning strategies \cite{GraphSAGE} and etc. to our method. Generally, this strategy consists of two
steps: aggregation and update. The aggregation step gathers the information from surrounding nodes and the update step transforms the node representations based on the aggregated information with a learnable weight matrix.


For each node, we simply average the representations of surrounding nodes and employ linear transformation to update the current node representations. Since both of the steps are independent of specific graph structures, our method can be applied even if node adjacency and the number of nodes vary. 
Denote the learnable linear transformation matrix as $\mathbf{W_{G}} \in \mathbb{R}^{C \times C}$. The message passing process can be formulated as 
\begin{equation}
    \begin{aligned}
        \hat{\mathbf{R}}_{\rm {G}}=\left ( \frac{1}{\mathbf{N_A}}\circ(\mathbf{A}\mathbf{R}_{\rm G}) \right ) \mathbf{W}_{\rm G},
    \end{aligned}
\end{equation}
where $\mathbf{N_A} = [\sum^{M}_{i=1} \mathbf{A}(1, i);...;\sum^{M}_{i=1} \mathbf{A}(M,i)]$ is the summation of each row of $\mathbf{A}$. 

To mitigate the over-smoothing effect, we employ a regularization on $\hat{\mathbf{R}}$ following \cite{ViG}, which is defined as
\begin{equation}
\begin{aligned}    \mathbf{R}_{\rm G}=\sigma(\hat{\mathbf{R}}_{\rm {G}}\mathbf{W_{\alpha}})\mathbf{W_{\beta}} + \hat{\mathbf{R}}_{\rm {G}},
\end{aligned}
\label{eq:message_pass}
\end{equation}
where $\mathbf{W_{\alpha}}, \mathbf{W_{\beta}}$ are two learnable linear matrics and $\sigma$ is an activation function (\eg, GeLu).

\textit{Note that this process is only a prototype version to demonstrate the effectiveness of the proposed hierarchical region-aware graph. It can be upgraded with other advanced graph reasoning approaches, which may result in additional improvement.}

After graph reasoning, we remap the updated representations back to the feature elements for feature enhancement. The essence is to revert Eq.~\eqref{eq:center_f}. For the feature $f^i$, the inverse mapping can be defined as
\begin{equation}
    \begin{aligned}
        \mathbf{F}^{i}_{\rm G}=\mathbf{D}^{i}\mathbf{R}^i, \quad \mathbf{R}^i = \mathbf{R}^{i}_{\rm G} {\mathbf{W}^{i}_{\rm inv}},
    \end{aligned}
    \label{eq:remap}
\end{equation}
where $\mathbf{R}_{\rm G}^i$ denotes the updated representations of graph nodes at $i$-th feature layer, $\mathbf{W}^{i}_{\rm inv}$ is a linear matrix to match the channel number of $f^i$, $\mathbf{D}^{i}$ is the feature element-region association matrix at $i$-th feature layer.
Then we integrate $\mathbf{F}^{i}_{\rm G}$ with the features $f^i$ as

\begin{equation}
    \begin{aligned}
        & f^{i}_{\rm G} \leftarrow \text{Reshape}(\mathbf{F}^{i}_{\rm G}),  f^{i} = \mu f^{i}_{\rm G} + f^{i},
    \end{aligned}
    \label{eq:mapping_fuse}
\end{equation}
where $\mu$ is a learnable parameter that controls the feature integration.

\subsection{Objective Functions}
\label{sec:training}

Note that the proposed method does not require extra supervision. Thus we only employ a task-specific loss function for training. 
Considering that the area of the manipulated regions in images is usually less than the authentic surroundings, we employ focal loss \cite{focal} for balancing the importance of authentic and manipulated pixels, which is defined as
\begin{equation}
\begin{aligned}
    \mathbf{L}(\bm{y}, \hat{\bm{y}}) = - \sum & \left ( \alpha \cdot (1 - \bm{y})^{\gamma} \cdot \hat{\bm{y}} \log (\bm{y}) \right. \\
    & \left.  + (1 - \alpha) \bm{y}^{\gamma} (1 - \hat{\bm{y}}) \log (1 - \bm{y}) \right ),
\end{aligned}
\end{equation}
where $\hat{\bm{y}}$ represents the ground truth mask, $ \alpha $ is the weight for balancing, and $ \gamma $ is the difficulty adjustment factor. The larger value of $ \gamma $ indicates a focus on more challenging samples. 

\section{Experiments}
\subsection{Experimental Settings}
\smallskip
\noindent{\bf Datasets.}
Following previous methods, we evaluate our method on five datasets, CASIA\cite{casia}, Coverage\cite{coverage}, NIST16\cite{nist16}, Columbia\cite{columbia}, and IMD20\cite{imd20}. 
The CASIA dataset contains 921 images in CASIAv1 and 5123 images in CASIAv2. We use CASIAv2 for training and CASIAv1 for testing. 
The coverage dataset comprises 100 images, with 75 images for training and 25 for testing. NIST16 has 564 images, where 404 images for training and 160 images for testing. The Columbia dataset consists of 180 images, and {IMD20} dataset consists of 2010 images. Note that these two datasets do not provide training sets. Thus we only use them for testing. 

\smallskip
\noindent{\bf Evaluation Metrics.}
Following previous work \cite{PSCC,ObjectFormer,ERMPC,HiFi}, we utilize the widely used F1 score for evaluation. The F1 score is calculated using the Equal Error Rate (EER) as the threshold. Moreover, we also employ pixel-level Area Under Curve (AUC) as complementary.




\smallskip
\noindent{\bf Implementation Details.}
Our method is implemented by PyTorch 11.3 with a Nvidia 3090 GPU. In the training phase, we utilize the AdamW optimizer with a learning rate of $6 \times 10^{-5}$ and a weight decay of $0.05$. We employ a linear warm-up learning strategy, where the learning rate is linearly increased during warm-up and then linearly decreased based on the iteration until it reaches $0$. The initial learning rate for warm-up is set to $1 \times 10^{-6}$, and warm-up is completed within $1.5k$ iterations. The training image size is set to $512 \times 512$ and the training iterations are set to $160k$. For feature integration, each layer's $\mu_i$ is learned separately with initialization of $1$. In the objective function, we set the hyperparameters as follows: $\alpha = 0.5, \gamma = 2$. We employ a plain feature pyramid detector \cite{UPerNet} as the decoder. The detection performance can be further improved if a more advanced decoder is employed.

\begin{table}[!t]
\centering
\small
\caption{\small Comparison of different methods. The \textbf{best} result is highlighted in bold while the \underline{second-best} result is marked by underlining.}
\vspace{-0.3cm}
\tabcolsep=0.4em
\resizebox{\linewidth}{!}{
\begin{tabular}{c|cccccc|cc}
\hline
\multirow{2}{*}{Method} & \multicolumn{2}{c}{CASIA} & \multicolumn{2}{c}{Coverage} & \multicolumn{2}{c|}{NIST16} & \multicolumn{2}{c}{avg} \\ 
    ~        & AUC  & F1   & AUC  & F1   & AUC  & F1 &  AUC & F1 \\ 
\hline
HGCN \cite{HGCN}   & - & 40.8 & - & 39.3 &  - & 71.0 &  -  & 50.4 \\
PSCC-Net \cite{PSCC}    & 87.5 & 55.4 & 94.1 & 72.3 & 99.1 & 74.2 & 93.7 & 69.8 \\

ObjFormer \cite{ObjectFormer} & 88.2 & 57.9 & 95.7 & 75.8 & 99.6 & 82.4 & 94.5 & 72.0 \\
EVP* \cite{EVP} & 86.2 & 63.6 & - & - & - & - & 86.2 & 63.6 \\
PCL \cite{PCL} & 75.1 & 46.7 & 91.7 & 62.0 & 94.6 & 78.0 & 87.1 & 62.2 \\
NCL* \cite{NCL} & 86.4 & 59.8 & 92.8 & 80.1 & 91.2 & 83.1 & 90.1 & 74.3 \\ 
HiFi-Net \cite{HiFi}    & 88.5 & 61.6 & \underline{96.1} & \underline{80.1} & 98.9 & \underline{85.0} & 94.6 & 75.5 \\

ERMPC  \cite{ERMPC}      & 90.4 & 58.6 & \textbf{98.4} & 77.3 & \textbf{99.7} & 83.6 & \textbf{96.1} & 73.1 \\
ACBG. \cite{ACBG} &  - & 66.9 & - & 71.1 & - & \textbf{93.3} & - & 77.1 \\
\hline
PSCC-Net$^\dagger$ \cite{PSCC}  & 85.2 & 53.6 & 91.1 & 71.6 & 96.4 & 70.1 & 90.3 & 65.1 \\
HiFi-Net$^\dagger$ \cite{HiFi}    & 85.9 & 59.0 & 86.2 & 63.4 & 87.4 & 62.8 & 86.5 & 61.7 \\
CAT-Net$^\dagger$ \cite{CAT-Net} & 77.3 & 40.6 & 79.2 & 47.8 & 89.2 & 61.5 & 81.9 & 50.0 \\ 
MVSS-Net$^\dagger$ \cite{MVSS} & 86.4 & 60.2 & 92.7 & 76.3 & 96.1 & 69.3 & 91.7 & 68.6 \\ 
ObjFormer$^\dagger$ \cite{ObjectFormer} & 71.9 & 42.6 & 68.4 & 41.3 & 91.4 & 68.2 & 77.2 & 50.7 \\ 
TruFor$^\dagger$ \cite{TruFor} & 92.4 & \underline{67.2} & 96.2 & 79.0 & 98.6 & 82.6 & 95.7 & 76.2 \\ 
\hline
Res-50  \cite{resnet}       & 87.0 & 54.0 & 90.5 & 69.8 & 97.1 & 64.7 & 91.5 & 62.8 \\
+ HRGR         & 88.5 & 56.2 & 92.7 & 71.3 & 97.2 & 64.1 & 92.8 & 63.9 \\
\hline
HRNet  \cite{hrnet}        & 90.1 & 62.2  & 90.3 & 68.6 & 97.4 & 66.5 & 92.6 & 65.8 \\
+ HRGR         & 91.4 & 64.2  & 90.9 & 69.5 & 98.4 & 67.1 & 93.6 & 66.9 \\
\hline 
EffNet-b4 \cite{efficient}      & 92.5 & 66.2 & 93.3 & 76.1 & 97.5 & 61.7 & 94.4 & 68.0 \\
+ HRGR         & \underline{92.9} & 66.7 & 93.9 & 74.8 & 97.2 & 63.4 & 94.7 & 68.3 \\
\hline 
Intern-tiny \cite{intern}   & 91.4 & 66.0 & 94.5 & 78.7 & 99.3 & 79.3 & 95.1 & 74.7 \\
+ HRGR         & \textbf{93.4} & \textbf{69.9} & 94.8 & \textbf{82.1} & 99.3 & 84.6 & \underline{95.8} & \textbf{78.9} \\
\hline
\end{tabular}}
\label{tab:finetune}
\end{table}

\begin{figure}[!t]
    \centering
    \includegraphics[width=1\linewidth]{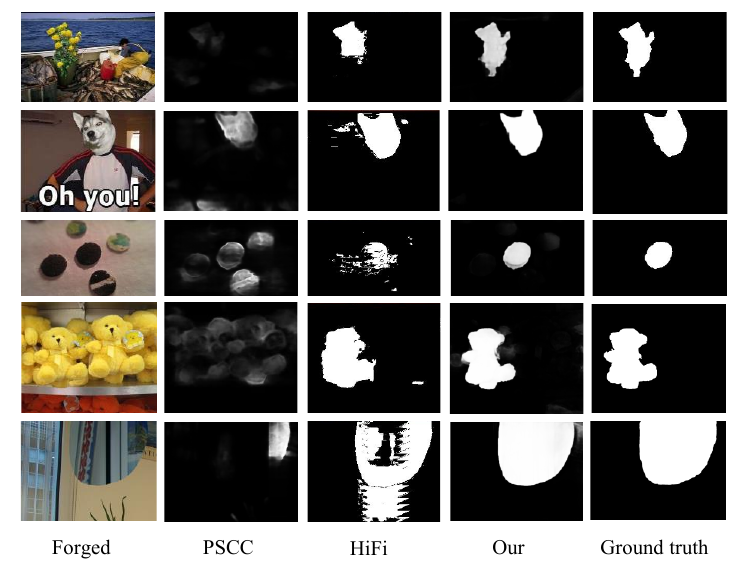}
    \vspace{-0.6cm}
    \caption{\small Qualitative visualization on CASIA, IMD20, NIST16, Coverage, and Columbia datasets (from top to bottom).}
    \label{fig:visualization}
    \vspace{-0.5cm}
\end{figure}

\subsection{Comparing with State-of-the-arts}
Following previous works \cite{PSCC,HiFi}, we explore the Fine-tuned scenario, that firstly pre-trains the methods using synthetic dataset in \cite{PSCC} and fine-tunes the pre-trained model on the training set of each dataset. Then we evaluate the performance on its testing set. This scenario represents the capacity to capture manipulation traces in a specific dataset. 

Following previous methods, the comparison is conducted on CASIA, Coverage, and NIST16 datasets since they provide training sets. 
Table \ref{tab:finetune} shows the results of our method compared to other state-of-the-art methods and graph-based methods under the fine-tuned scenario. 
We would like to clarify that EVP and NCL were trained without using synthetic dataset as others, while TransF. was tested on the Coverage dataset without fine-tuning. These three methods are marked by $(*)$. 
Moreover, the methods marked with $(\dagger)$  are the ones that we have reproduced. Specifically, PSCC-Net and HiFi-Net were reproduced using the code provided by the respective authors, while the remaining methods were reproduced with IMDLBenCo \cite{benco}. 

We can observe that our method improves the performance on these base networks by $0.8\%$ in AUC and $1.7\%$ in the F1 score on average, demonstrating the efficacy of our method to improve the capacity of learning specific manipulation traces. Compared to state-of-the-arts, our method achieves an average F1 score of $78.9\%$ across the three datasets, surpassing the highest-performing HiFi-Net by $3.3\%$. Particularly on the CASIA dataset, our method outperforms HiFi-Net with a $3.0\%$ improvement in the AUC score and a $8.3\%$ increase in the F1 score. These results demonstrate the superiority of the proposed method in capturing subtle manipulation traces. Moreover, our method only introduces a few additional parameters (1.45M) and FLOPS (1.49G). Fig.~\ref{fig:visualization} shows several visualizations of different methods on five datasets.

\subsection{Analysis} 

\smallskip
\noindent{\bf Ablation Study.} 
This part investigates the effect of different settings: 1) \textbf{Baseline} indicates solely using Intern-tiny network. 2) \textbf{Feature region $\rightarrow$ Grid} indicates replacing the feature partition module with the original regular grid. 3) \textbf{Graph reasoning $\rightarrow$ Linear} refers to replacing graph reasoning with several linear layers (equivalent to fully connected graph reasoning). 4) \textbf{Graph reasoning $\rightarrow$ MHSA} refers to replacing graph reasoning with several Mutil-Head Self Attention layers. 5) \textbf{Hierarchical $\rightarrow$ Layer-wise} means replacing the hierarchical graph modeling with layer-wise independent graph modeling and performing graph reasoning. 6) \textbf{DFP w/o x- and y- coords} means that we do not additionally concatenate the x and y coordinates in the clustering process of the DFP module.
Table \ref{tab:ablation} displays the results of ablation experiments. It can be observed that when any module is replaced, the performance drops significantly, \eg, F1 scores decreasing by $2.9\%$, $2.6\%$, $3.1\%$, $1.9\%$, $1.0\%$ in the finetune stage. Nevertheless, only replacing one component still has an improvement compared to the baseline, which confirms the effectiveness of each component.

\begin{table}[!t]
\centering
\small
\caption{\small Ablation study of different settings on CASIA dataset.}
\vspace{-0.3cm}
\begin{tabular}{ccccc}
\hline
\multirow{2}{*}{Method} & \multicolumn{2}{c}{w/o Fine-tune} & \multicolumn{2}{c}{w/ Fine-tune} \\ 
    ~             & AUC  & F1   & AUC  & F1  \\ 
\hline
Baseline          & 84.1 & 48.3 & 91.4 & 66.0 \\
Feature region $\rightarrow$ Grid & 85.1 & 49.6 & 92.1 & 67.0 \\
Graph reasoning $\rightarrow$ Linear & 85.8 & 50.8 & 92.1 & 67.3 \\
Graph reasoning $\rightarrow$ MHSA & 85.3 & 49.1 & 91.7 & 66.8 \\
Hierarchical $\rightarrow$ Layer-wise & 85.6 & 50.6 & 92.5 & 68.0 \\
DFP w/o x- and y- coords  & 85.4 & 49.5 & 93.3 & 68.9 \\
Full & \textbf{86.3} & \textbf{52.4} & \textbf{93.4} & \textbf{69.9}   \\
\hline
\end{tabular}
\label{tab:ablation}
\vspace{-0.3cm}
\end{table}

\smallskip
\noindent{\bf Performance without Fine-tune.} 
This scenario represents that the methods are pre-trained on synthetic images and directly tested on other datasets, aiming to exhibit the generalizability of detection. Table~\ref{tab:pretrained-ours} shows the performance of the different base networks before and after using our method HRGR. These methods are pre-trained on the synthetic dataset from \cite{PSCC} and directly tested on all datasets. 
For a fair comparison, we use the same configuration for each experimental group. It can be seen that the performance of base networks generally improves after using our method. 

\begin{table}[!t]
\centering
\small
\caption{\small Performance without fine-tune.}
\vspace{-0.3cm}
\tabcolsep=0.4em
\resizebox{\linewidth}{!}{
\begin{tabular}{c|cccccccccc|cc}
\hline
\multirow{2}{*}{Base network} & \multicolumn{2}{c}{CASIA} & \multicolumn{2}{c}{Coverage} & \multicolumn{2}{c}{NIST16} & 
\multicolumn{2}{c}{Columbia} & \multicolumn{2}{c|}{IMD20} & \multicolumn{2}{c}{avg} \\ 
    ~        & AUC  & F1   & AUC  & F1   & AUC  & F1 &  AUC & F1 & AUC  & F1 &  AUC & F1 \\ 
\hline
ResNet-50     & 82.0 & 41.2 & 78.3 & 42.5 & 80.6 & 28.1 & \textbf{92.8} & \textbf{80.3} & 78.6 & 26.3 & 82.4 & 43.6 \\
+ HRGR        & \textbf{82.3} & \textbf{42.6} & \textbf{78.9 }& \textbf{43.0} & \textbf{80.7} & \textbf{29.0} & 92.1 & 80.1 & \textbf{79.2} & \textbf{27.1} & \textbf{82.6} & \textbf{44.3} \\
\hline
HRNet         & 82.7 & 44.9 & 80.3 & 45.6 & 81.7 & 31.5 & 95.1 & 87.2 & 79.6 & 28.4 & 83.8 & 47.5 \\
+ HRGR        & \textbf{84.0} & \textbf{46.3} & \textbf{81.0} & \textbf{46.0} & \textbf{82.3} & \textbf{34.6} & \textbf{95.7} & \textbf{88.2} & \textbf{80.4} & \textbf{29.9} & \textbf{84.6} & \textbf{49.0} \\
\hline
EffNet-b4     & 82.1 & 43.3 & 74.3 & 35.6 & \textbf{83.2} & 30.1 & 85.8 & 68.5 & \textbf{81.3} & 31.5 & 81.3 & 41.8 \\
+ HRGR        & \textbf{83.5} & \textbf{44.9} & \textbf{75.8} & \textbf{37.8} & 81.9 & \textbf{32.2} & \textbf{86.4} & \textbf{68.7} & 80.9 & \textbf{31.7} & \textbf{81.7} & \textbf{43.0} \\
\hline
Intern-tiny   & 84.1 & 48.3 & 80.3 & 49.3 & \textbf{86.6} & 39.3 & 94.7 & 88.8 & 79.5 & 29.5 & 85.0 & 51.0 \\ 
+ HRGR        & \textbf{86.3} & \textbf{52.4} & \textbf{83.9} & \textbf{54.7} & 86.2 & \textbf{39.4} & \textbf{96.4} & \textbf{91.2} & \textbf{79.8} & \textbf{30.4} & \textbf{86.5} & \textbf{53.6} \\
\hline
\end{tabular}}
\label{tab:pretrained-ours}
\vspace{-0.3cm}
\end{table}

\begin{figure}[!t]
    \centering
    \includegraphics[width=0.9\linewidth]{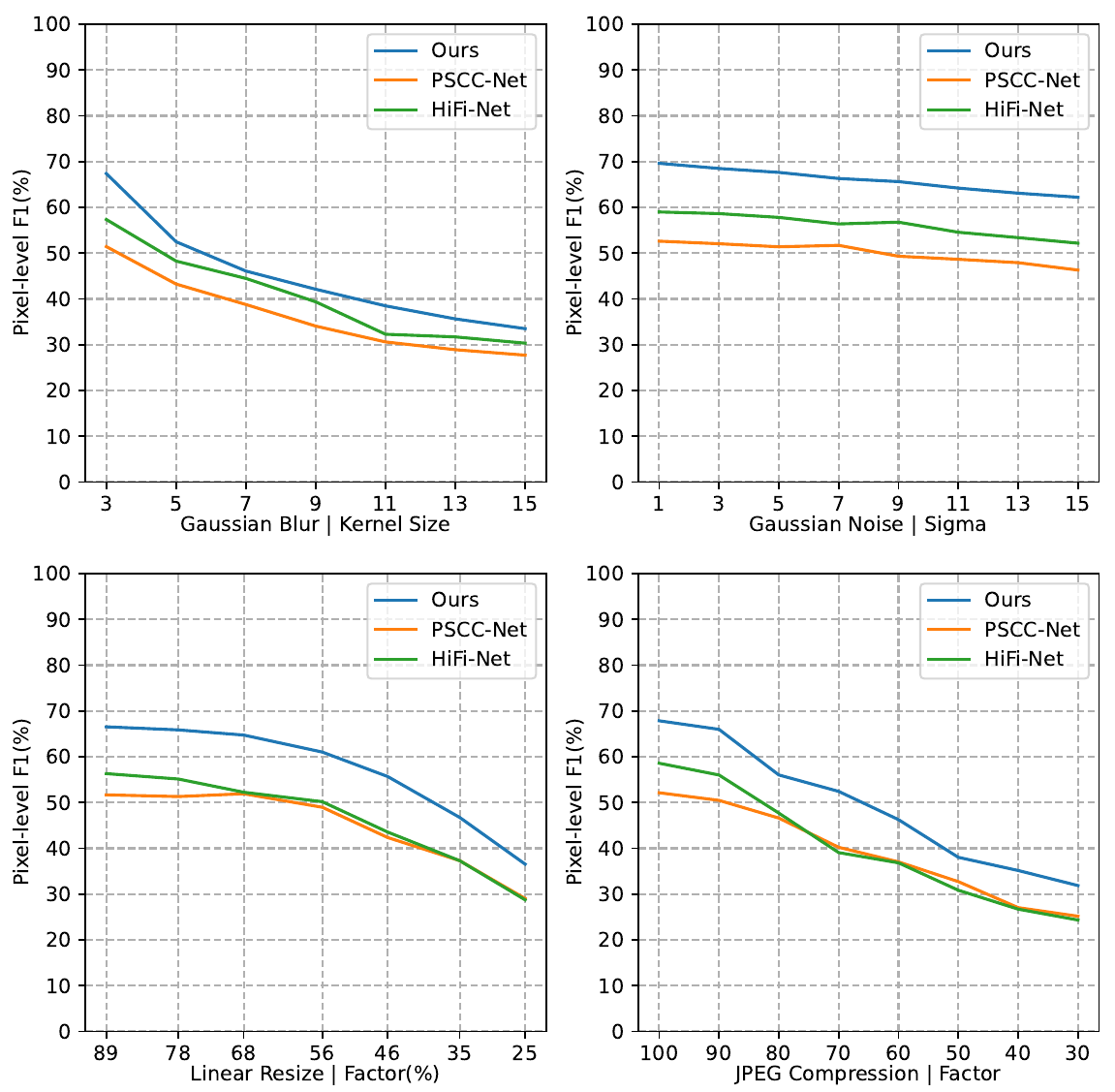}
    \vspace{-0.3cm}
    \caption{\small Robustness analysis under various perturbations on CASIA dataset.}
    \label{fig:robust}
    \vspace{-0.4cm}
\end{figure}

\smallskip
\noindent{\bf Robustness.}
To evaluate the robustness of our method, we follow the perturbation settings in \cite{PSCC}, which degrades manipulated images obtained from CASIA under different preprocesses, including resizing images with different scales, Gaussian Blurring with kernel size, adding Gaussian Noise with standard deviation ``Sigma'', and performing JPEG Compression with quality factor. The evaluation results in comparison to PSCC-Net and HiFi-Net are shown in Fig.~\ref{fig:robust}, which reveals that our method can resist certain perturbations. 

\section{Conclusion}
This paper introduces a new method, Hierarchical Region-aware Graph Reasoning (HRGR), to enhance image manipulation detection. Specifically, we describe a differentiable feature partition strategy to formulate feature regions and construct a hierarchical region-aware graph on these regions. Then we introduce a structure-agnostic graph reasoning to improve the feature effectiveness. These processes are iteratively updated during training, leading to mutual boosting. Extensive experiments demonstrate the effectiveness of our method and its potential as a plug-and-play component for classical models. 

{\small
\bibliographystyle{IEEEbib}
\bibliography{icme2025references}

\begin{thebibliography}{10}

\bibitem{recent}
Rahul Thakur and Rajesh Rohilla,
\newblock ``Recent advances in digital image manipulation detection techniques: A brief review,''
\newblock {\em Forensic science international}, 2020.

\bibitem{ERMPC}
Dong Li, Jiaying Zhu, Menglu Wang, Jiawei Liu, Xueyang Fu, and Zheng-Jun Zha,
\newblock ``Edge-aware regional message passing controller for image forgery localization,''
\newblock in {\em CVPR}, 2023.

\bibitem{PSCC}
Xiaohong Liu, Yaojie Liu, Jun Chen, and Xiaoming Liu,
\newblock ``Pscc-net: Progressive spatio-channel correlation network for image manipulation detection and localization.,''
\newblock {\em TCSTV}, 2022.

\bibitem{ObjectFormer}
Junke Wang, Zuxuan Wu, Jingjing Chen, Xintong Han, Abhinav Shrivastava, Ser-Nam Lim, and Yu-Gang Jiang,
\newblock ``Objectformer for image manipulation detection and localization,''
\newblock in {\em CVPR}, 2022.

\bibitem{CAT-Net}
Myung-Joon Kwon, Seung-Hun Nam, In-Jae Yu, Heung-Kyu Lee, and Changick Kim,
\newblock ``Learning jpeg compression artifacts for image manipulation detection and localization,''
\newblock {\em IJCV}, 2022.

\bibitem{ACBG}
Wenxi Liu, Hao Zhang, Xinyang Lin, Qing Zhang, Qi~Li, Xiaoxiang Liu, and Ying Cao,
\newblock ``Attentive and contrastive image manipulation localization with boundary guidance,''
\newblock {\em TIFS}, 2024.

\bibitem{TruFor}
Fabrizio Guillaro, Davide Cozzolino, Avneesh Sud, Nicholas Dufour, and Luisa Verdoliva,
\newblock ``Trufor: Leveraging all-round clues for trustworthy image forgery detection and localization,''
\newblock in {\em CVPR}, 2023.

\bibitem{HGCN}
Wenyan Pan, Zhili Zhou, Miaogen Ling, Xin Geng, and QM~Wu,
\newblock ``Learning hierarchical graph representation for image manipulation detection,''
\newblock {\em arXiv preprint arXiv:2201.05730}, 2022.

\bibitem{SpixelFCN}
Fengting Yang, Qian Sun, Hailin Jin, and Zihan Zhou,
\newblock ``Superpixel segmentation with fully convolutional networks,''
\newblock in {\em CVPR}, 2020.

\bibitem{SSN}
Varun Jampani, Deqing Sun, Ming-Yu Liu, Ming-Hsuan Yang, and Jan Kautz,
\newblock ``Superpixel sampling networks,''
\newblock in {\em ECCV}, 2018.

\bibitem{GraphSAGE}
Will Hamilton, Zhitao Ying, and Jure Leskovec,
\newblock ``Inductive representation learning on large graphs,''
\newblock {\em Advances in neural information processing systems}, 2017.

\bibitem{ViG}
Kai Han, Yunhe Wang, Jianyuan Guo, Yehui Tang, and Enhua Wu,
\newblock ``Vision gnn: An image is worth graph of nodes,''
\newblock {\em Advances in Neural Information Processing Systems}, 2022.

\bibitem{focal}
Tsung-Yi Lin, Priya Goyal, Ross Girshick, Kaiming He, and Piotr Doll{\'a}r,
\newblock ``Focal loss for dense object detection,''
\newblock in {\em ICCV}, 2017.

\bibitem{casia}
Jing Dong, Wei Wang, and Tieniu Tan,
\newblock ``Casia image tampering detection evaluation database,''
\newblock in {\em ChinaSIP}, 2013.

\bibitem{coverage}
Bihan Wen, Ye~Zhu, Ramanathan Subramanian, Tian-Tsong Ng, Xuanjing Shen, and Stefan Winkler,
\newblock ``Coverage — a novel database for copy-move forgery detection,''
\newblock in {\em ICIP}, 2016.

\bibitem{nist16}
HY~Guan, YY~Lee, A~Yates, A~Delgado, D~Zhou, D~Joy, and A~Pereira,
\newblock ``Nist nimble 2016 datasets,'' 2016.

\bibitem{columbia}
Tian-Tsong Ng, Jessie Hsu, and Shih-Fu Chang,
\newblock ``Columbia image splicing detection evaluation dataset,''
\newblock {\em DVMM lab. Columbia Univ CalPhotos Digit Libr}, 2009.

\bibitem{imd20}
Adam Novozamsky, Babak Mahdian, and Stanislav Saic,
\newblock ``Imd2020: A large-scale annotated dataset tailored for detecting manipulated images,''
\newblock in {\em WACVW}, 2020.

\bibitem{HiFi}
Xiao Guo, Xiaohong Liu, Zhiyuan Ren, Steven Grosz, Iacopo Masi, and Xiaoming Liu,
\newblock ``Hierarchical fine-grained image forgery detection and localization,''
\newblock in {\em CVPR}, 2023.

\bibitem{UPerNet}
Tete Xiao, Yingcheng Liu, Bolei Zhou, Yuning Jiang, and Jian Sun,
\newblock ``Unified perceptual parsing for scene understanding,''
\newblock in {\em ECCV}, 2018.

\bibitem{EVP}
Weihuang Liu, Xi~Shen, Chi-Man Pun, and Xiaodong Cun,
\newblock ``Explicit visual prompting for low-level structure segmentations,''
\newblock in {\em CVPR}, 2023.

\bibitem{PCL}
Yuyuan Zeng, Bowen Zhao, Shanzhao Qiu, Tao Dai, and Shu-Tao Xia,
\newblock ``Towards effective image manipulation detection with proposal contrastive learning,''
\newblock {\em TCSVT}, 2023.

\bibitem{NCL}
Jizhe Zhou, Xiaochen Ma, Xia Du, Ahmed~Y Alhammadi, and Wentao Feng,
\newblock ``Pre-training-free image manipulation localization through non-mutually exclusive contrastive learning,''
\newblock in {\em ICCV}, 2023.

\bibitem{MVSS}
Chengbo Dong, Xinru Chen, Ruohan Hu, Juan Cao, and Xirong Li,
\newblock ``Mvss-net: Multi-view multi-scale supervised networks for image manipulation detection,''
\newblock {\em TPAMI}, 2022.

\bibitem{resnet}
Kaiming He, Xiangyu Zhang, Shaoqing Ren, and Jian Sun,
\newblock ``Deep residual learning for image recognition,''
\newblock in {\em CVPR}, 2016.

\bibitem{hrnet}
Jingdong Wang, Ke~Sun, Tianheng Cheng, Borui Jiang, Chaorui Deng, Yang Zhao, Dong Liu, Yadong Mu, Mingkui Tan, Xinggang Wang, et~al.,
\newblock ``Deep high-resolution representation learning for visual recognition,''
\newblock {\em TPAMI}, 2020.

\bibitem{efficient}
Mingxing Tan and Quoc Le,
\newblock ``Efficientnet: Rethinking model scaling for convolutional neural networks,''
\newblock in {\em ICML}, 2019.

\bibitem{intern}
Wenhai Wang, Jifeng Dai, Zhe Chen, Zhenhang Huang, Zhiqi Li, Xizhou Zhu, Xiaowei Hu, Tong Lu, Lewei Lu, and Hongsheng Li,
\newblock ``Internimage: Exploring large-scale vision foundation models with deformable convolutions,''
\newblock in {\em CVPR}, 2023.

\bibitem{benco}
Xiaochen Ma, Xuekang Zhu, Lei Su, Bo~Du, Zhuohang Jiang, Bingkui Tong, Zeyu Lei, Xinyu Yang, Chi-Man Pun, Jiancheng Lv, and Jizhe Zhou,
\newblock ``Imdl-benco: A comprehensive benchmark and codebase for image manipulation detection and localization,''
\newblock in {\em NeurIPS}, 2024.

\end{thebibliography}
}

\end{document}